# Build a Compact Binary Neural Network through Bit-level Sensitivity and Data Pruning

**Yixing Li** and **Fengbo Ren**
School of Computing, Informatics, and Decision Systems Engineering
Arizona State University
{yixingli, renfengbo}@asu.edu

## Abstract

Convolutional neural network (CNN) has been widely used for vision-based tasks. Due to the high computational complexity and memory storage requirement, it is hard to directly deploy a full-precision CNN on embedded devices. The hardware-friendly designs are needed for resource-limited and energy-constrained embedded devices. Emerging solutions are adopted for the neural network compression, e.g., binary/ternary weight network, pruned network and quantized network. Among them, Binarized Neural Network (BNN) is believed to be the most hardware-friendly framework due to its small network size and low computational complexity. No existing work has further shrunk the size of BNN. In this work, we explore the redundancy in BNN and build a compact BNN (CBNN) based on the bit-level sensitivity analysis and bit-level data pruning. The input data is converted to a high dimensional bit-sliced format. In post-training stage, we analyze the impact of different bit slices to the accuracy. By pruning the redundant input bit slices and shrinking the network size, we are able to build a more compact BNN. Our result shows that we can further scale down the network size of the BNN up to 3.9x with no more than 1% accuracy drop. The actual runtime can be reduced up to 2x and 9.9x compared with the baseline BNN and its full-precision counterpart, respectively.

## 1 Introduction

Vision-based applications can be found in many embedded devices for classification, recognition, detection and tracking tasks [Tian *et al.*, 2015; Hu *et al.*, 2015]. Specifically, convolutional neural network (CNN) has become the core architecture for those vision-based tasks [LeCun *et al.*, 2015]. Since it can outperform conventional feature selection-based algorithm in terms of accuracy, it becomes more and more popular. Advanced driver-assistance system (ADAS) can either used CNNs for guiding autonomous driving or alerting the driver of predicted risk [Tian *et al.*, 2015]. It is obvious that ADAS depends on a low-latency system to get a timely reaction. Artificial intelligence (AI) applications also explode in smartphones, such as automatically tagging the photos, face detection and so on [Schroff *et al.*, 2015; Hu *et al.*, 2015]. Apple has been reported working on an "Apple Neural Engine" to partially move their AI processing module on device [Lumb *et al.*, 2017]. If processing the users' requests through sending them to the data center, there will be much overhead of the latency and power consumption caused by the commutation. As such, on-device AI processing is the future trend to balance power efficiency and latency. However, CNNs are known to have high computation complexity, which makes it hard to directly deploy on embedded devices. Therefore, compressed CNNs are in demand.

In the early stage, research work of hardware-friendly CNNs have focused on reducing the network precision down to 8-16 bits in the post-training stage [Suda *et al.*, 2016], which either has a limited reduction or suffers from severer accuracy drop. Lately, in-training techniques have been brought up, achieving much higher compression ratio. BinaryConnect, BNN, TernaryNet and XNOR-Net [Courbariaux *et al.*, 2015; Hubara *et al.*, 2016; Rastegari *et al.*, 2016; Zhu *et al.*, 2016] have pushed to reduce the weight to binary or ternary (-1, 0, +1) values. Network pruning [Han *et al.*, 2015] reduces the network size (the memory size for all the parameters) by means of reducing the number of connections. Regarding the network size, pruned network and TernaryNet can achieve 13x and 16x reduction [Zhu *et al.*, 2016; Han *et al.*, 2015], respectively. While BinaryConnect, BNN and XNOR-Net can achieve up to 32x reduction. In terms of computational complexity, only BNN and XNOR-Net with both binarized weights and activations can simply replace convolution operation with bitwise XNOR and bit count operation. However, XNOR-Net has additional scaling factor filters in each layer, which brings overhead to both memory and computation cost. From the above, BNN is the optimal solution for hardware deployment when considering both network size and computational complexity. In addition, BNN has drawn a lot of attention in hardware community [Li *et al.*, 2017]. However, no existing study has explored scaling down BNN for a more efficient inference stage.

This work is the first one that explores and proves that there is still redundancy in BNN. The proposed flow to reduce the network size is triggered by conversion and analysis of input data rather than the network body, which is rarely seen in previous work. A novel flow is proposed to prune out the redundant input bit slices and rebuild a compact BNN through bit-level sensitivity analysis. Experiment results show that the compression ratio of the network size is achieving up to 3.9x with no more than 1% accuracy drop.

The rest of the paper is organized as follows. Section 2 discusses the related work for network compression and explains why BNN is a more superior solution be deployed on the hardware. Section 3 validates the hypothesis that BNN has redundancy and proposes a novel flow to build a compact BNN. Experiment results and discussion are shown in Section 4. Section 5 concludes the paper.

## 2 Related work

Regarded to hardware-friendly oriented designs, it is not fair only emphasize compressing the network size. Other than that, the computational complexity is also essential. In this section, we first discuss and evaluate the related work for network compression by emphasizing both factors. We also present a simple benchmark study to help the reader better understand the computational complexity in terms of hardware resource of existing work. It can reveal why BNN is a more superior solution to be deployed on the hardware.

BinaryConnect [Courbariaux *et al.*, 2015] is a study in the early stage of exploring the binarized weight neural network. In the BinaryConnect network, the weights are binary values while the activations are still non-binary. Arbitrary value multiplies +1/-1 is equivalent to a conditional bitwise NOR operation. Hence, the convolution operation can be decomposed into conditional bitwise NOR operations and accumulation. It is a big step moving from full-precision multiplication to much simpler bitwise operations.

BNN [Hubara *et al.*, 2016] is the first one that builds a network with both binary weights and activations. The convolution operation has been further simplified as bitwise XNOR (Exclusive-NOR) and bit count operations. The hardware resource cost is minimized for GPU, FPGA and ASIC implementation. For GPU implementation, 32-bit bitwise XNOR can be implemented in a single clock cycle with one CUDA core. For FPGA and ASIC implement, there is no need to use DSP (Digital Signal Processor) resources anymore, which is relatively costly. Simple logic elements (LUTs, Look Up Tables) can be used to map bitwise XNOR and bit count operations, which makes it easy to map highly parallel computing engines to achieve high throughput and low latency.

XNOR-Net [Rastegari *et al.*, 2016] also builds the network based on binary weights and activations. However, it introduces a filter of full-precision scaling factors in each convolutional layer to ensure a better accuracy rate. Additional non-binary convolution operations are needed in each convolutional layer, which cost additional time and computing resources. Thus, the XNOR-Net is not strictly a fully binarized network.

TernaryNet [Zhu *et al.,* 2016] holds ternary (-1, 0, +1) weights for its network. By increasing the precision level of the weights, it enhances the accuracy rate. Since ternary weights have to be encoded in 2 bits, the computational complexity will at least double, compared with BinaryConnect.

Network pruning [Han *et al.*, 2015] is revealed as the most popular technique for compressing pre-trained full-precision or reduced-precision CNNs (weights of the reduced-precision CNN are usually in the range of 8 bit - 16 bit [Suda *et al.*, 2016]). It compresses the network by pruning out the useless weights, which gains speedup mainly by reducing the network size. Unlike all the other technique mentioned above, neither the weights nor activations of a pruned network are binary or ternary. Still, the computation complexity of the full-precision or reduced-precision multiply-add operation is much higher than that of the BNN.

We implement a $W_{(10,10)} \times A_{(10,10)}$ matrix multiplication on a Xilinx Virtex-7 FPGA board for analyzing the computational complexity of the different architecture that mentioned above. The precision of elements in $W$ and $A$ are the same as the precision of weights and activations in each architecture. The matrix multiplication is fully mapped onto the FPGA. In other words, we don't reuse any hardware resource. So the resource utilization is a good reflection of computational complexity. Since 16 bits are enough to maintain the same accuracy rate as the full precision network [Suda *et al.*, 2016], we set precision to 16 bits for any full precision weights or activations. For the pruned network, we set 84% of the $W$ in the pruned network as zero for a fair comparison. (Since pruned network can get up to 13x reduction [Han *et al.*, 2015] while BNN can get 32x, the size of the pruned network is 32/13=2.5x larger. With 16-bit weights, the total number of non-zero weights of the pruned network is 2.5/16=16% of that of the binarized weight cases.) As shown in Fig. 1, BNN apparently consumes the least amount of hardware resource among all these architecture.

In summary, for all the methods mentioned above, pruning can be categorized as connection reduction, while the rest can be categorized as precision reduction. However, both kinds of methods cannot be applied to the BNN. For pruning, it prunes the weights that are close to zero value. However, the weights of BNN are already constrained to -1/+1. For precision reduction, BNN has already reached the lower bound.

Since CNNs are believed to have huge redundancy, we hypothesize that BNN also has redundancy and it is able to get a more compact BNN. To our best knowledge, no existing work is inspired by analyzing and reducing the input precision. We are the 1[st] to explore the BNN redundancy by the bit-level analysis of the input data. We will validate our hypothesis step by step in the next section.

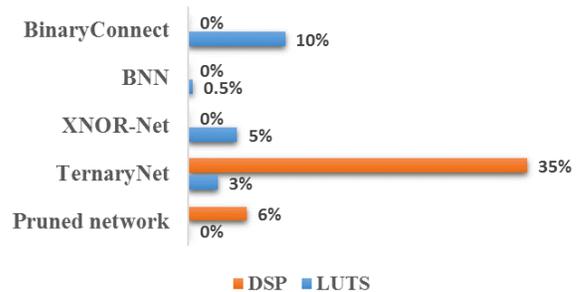

Figure 1. Resource consumption of $W_{(10,10)} \times A_{(10,10)}$ multiplication on a Xilinx Virtex-7 FPGA for different architecture

In the following paragraphs, BNN is referring to Hubara et al.'s work, a 9-layer binarized CNN, which is our baseline model. The reconstructed BNN and CBNN is referring to the reconstructed model we used for sensitivity analysis and the final compact BNN with shrunk network size, respectively.

## 3 Build a compact BNN

First, we need to reconstruct and train a new model for sensitivity analysis. Section 3.1 demonstrates the model reconstruction of BNN. Then, with the reconstructed BNN, Section 3.2 will prove the redundancy exists in BNN and decide the prunable bit slices through bit-level sensitivity analysis in post-training stage. Finally, Section 3.3 presents the guide to rebuilding a more compact BNN (CBNN) that triggered by input data pruning.

### 3.1 BNN reconstruction

In this section, we first illustrates reformatting the input and modifying the first layer for the BNN reconstruction. Then, the training method is presented.

**Bit-sliced binarized input**

All the existing binarized or ternarized neural networks take the non-binary format data as the network input, which means the computation of the first layer is the inner product of a non-binary matrix and a binary matrix. In all the other layers, the convolution operation can be implemented as the XNOR dot-product operation, which is simplified as bit count in GPU or XNOR logic in FPGA/ASIC. By contrast, the computation of the first layer is much more complicated. Intuitively, the bit-sliced input can enable XNOR dot-product operation in the first layer. This inspires us to explore the feasibility of converting the dataset into the bit-sliced format.

A single image in the dataset can be represented as $data_{(W,H,C)}$, where $W$ is the width, $H$ is the height, and $C$ is the number of channels, as shown in Fig. 2. The raw data is usually stored in the format of a non-negative integer with the maximum value of $A$. Then a lossless conversion from integer (fixed-point) to $N$-bit binary format is defined as the *int2b* function.

$$data^b_{(W,H,C')} = int2b(data_{(W,H,C)}, N), \quad (1)$$

where $C' = C \times N$ and $N = ceil(log_2(A+1))$. After *int2b* conversion, each channel of an image is expanded to $N$ channels in binary format.

**Non-binary first layer**

Experimental observation shows that the bit-sliced input has a negative impact on the accuracy rate. There are two main reasons. Since the input data is in the bit-sliced format, the data-preprocessing methods, e.g., mean removal, normalization, ZCA whitening, cannot be applied here, which results in an accuracy drop. In addition, compared with a standard first layer in BNN, the computational complexity drops, which may hurt the accuracy rate. Therefore, we assign the first layer with full-precision float-point weights to keep the computational complexity of the first layer the same as a standard first layer in BNN.

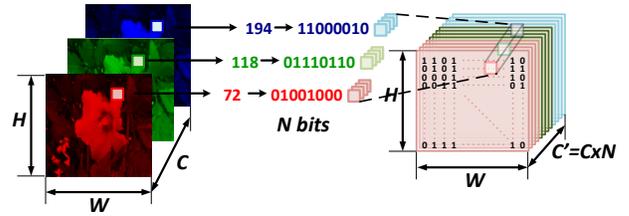

Figure 2. Conversion from fixed-point input to bit-sliced binary input

Although the network size increases, the growth is somewhat limited. For example, in a 9-layer BNN [Hubara *et al.*, 2016], the size of the first layer is only 0.02% of the entire network. It has been proved that 16-bit quantization for the weights is enough to preserve the accuracy [Suda *et al.*, 2016]. With the bit-slice input, the network size will slightly increase by 3%, which can be negligible.

With the bit-sliced input and non-binary first layer, we reconstruct the BNN model and refer it as the reconstructed BNN. Although the computational complexity is the same, the new structure helps to reduce the redundancy in BNN, which will be elaborated in the following sections.

**Binary constrained training**

We adopt the training method proposed by Hubara et al. [Hubara *et al.*, 2016]. The objective function is shown in Eq. 2, where $W_1$ represents the weights in the non-binary first layer and $W_l$ represents the weights in all the other binary layers. The loss function $L$ here is a hinge loss. In the training stage, the full-precision reference weights $W_l$ are used for the backward propagation, and the binarized weights $W_l^b = clip(W_l)$ [Hubara *et al.*, 2016] are used in the forward propagation. As Tang et al. propose in [Tang et al., 2017], the reference weights in the binary layers $W_l (l \geq 2)$ should be punished if they are not close to +1/-1. Also, a $L_2$ regularization term is applied for the non-binary first layer.

$$J(W_l, W_1, b) = L(W_l^b, W_1, b) + \lambda (avg \sum_{l=2}^{L}(1 - \|W_l\|_2^2) + avg(\|W_1\|_2^2)) \quad (2)$$

### 3.2 Sensitivity analysis

We use the training method in Section 3.1 to train a reconstructed BNN model with the bit-sliced input and non-binary first layer. In the post-training stage, we evaluate the sensitivity of the bit-sliced input to the accuracy performance.

As shown in Fig. 3, the reconstructed BNN is pre-trained as initial. Then, the $N^{th}$ bit ($N^{th}$ least significant bit) slices in RGB channels are substituted with binary random bit slices. The reason why we use binary random bit slices other than pruning is that, pruning will reduce the size of the network. We want to eliminate any other factors that can influence the accuracy performance. If the difference between the actual inference error $ERR_{inf}$ and the reference point $ERR_{ref}$ ($\Delta ERR = ERR_{inf} - ERR_{ref}$) is less than an error-tolerant threshold $ERR_{th}$, the $N^{th}$ bit slices are classified as prunable.

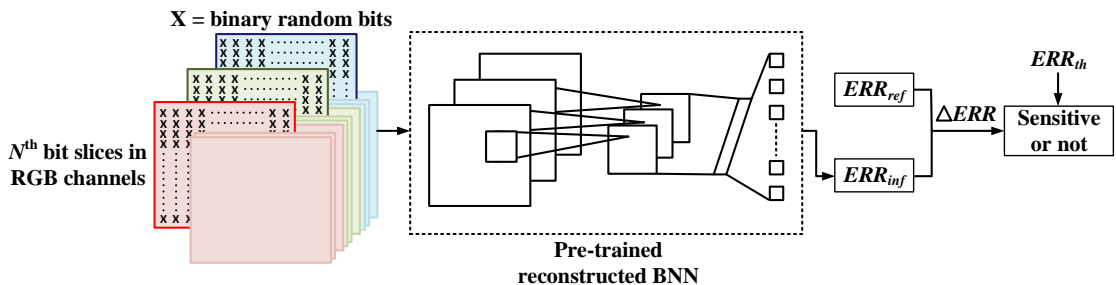
Figure 3. Sensitivity analysis of the reconstructed BNN with distorted input

Without retraining the network, the error brought by random bit slices will propagate throughout the entire network. Even so, there can be merely no accuracy drop in the inference stage. Then, it can be inferred that these bit slices with less sensitivity to the accuracy performance are useless in the training stage. It also indicates that the existing redundancy in BNN allows us to further shrink the network size. After evaluating the sensitivity of each bit slice, we can also analyze the sensitivity of a stack of slices by using the same method. Then we can find a collection of insensitive bit slices which are prunable in the training stage. If $P$ out of $N$ slices are categorized as accuracy insensitive, the number of channels $C'$ can be reduced by $N/P$ times. That is to say, the size of the input array is reduced by $N/P$ times.

### 3.3 Rebuild a compact network

In the most popular CNN architectures, such as AlexNet [Krizhevsky *et al.*, 2012], VGG [Simonyan and Zisserman, 2014] and ResNet [He *et al.*, 2016], the depth incremental ratio of feature map from one layer to the next layer is either keeping as 2x or remaining the same. Intuitively speaking, it is useful to keep the same depth incremental ratio across the entire network. Thus, a good starting point of rebuilding a compact BNN (CBNN) is shrinking the depth of all the layers by $N/P$ times. Since there is a quadratic relation between depth and the network size, the reduction of the network size of the CBNN is expected to be $(N/P)^2$ times.

Although we haven't explored how to build an accurate model to optimize the network compression ratio, we emphasize the entire flow (presented in Section 3) that reduces the redundancy of the BNN and enables speedup in the inference stage with the CBNN. In Section 4, we will present and discuss the performance results corresponding to each subsection in Section 3.

## 4 Result and discussion

We will first walk through the flow presented in Section 3 with experimental results on the Cifar-10 classification task in Section 4.1. Section 4.2 will present additional results on SNVH, Chars74K and GTSRB datasets. For the experiment setup, we build our model based upon Hubara et al.'s BNN in Theano. The description of each dataset is listed as follow.

**Cifar-10** [Krizhevsky and Hinton, 2009]. This is a dataset for 10-category classification task with 32×32 RGB images. The training dataset contains 50,000 images and the testing dataset contains 20,000.

**SVHN** (The Street View House Numbers) [Netzer *et al.*, 2011]. This dataset is a real-world house number dataset from Google Street View images. It has 73257 digits for training and 26032 digits for testing, with the image size of 32×32.

**Char74K** [Campos *et al.*, 2009]. This dataset contain 62 characters (0-9, A-Z and a-z) from both natural images and synthesized images. 80% of the Char74K images serve as the training set and the rest 20% serve as the testing set, with the image size of 56×56.

**GTSRB** (The German Traffic Sign Benchmark) [Stallkamp *et al.*, 2011]. This dataset includes 43-class traffic signs. We resize the traffic sign images to 32×32. It has 39209 training data and 12630 testing data.

### 4.1 Experiment on Cifar-10

Subsections of 4.1 show the experimental results corresponding to the methodology in Section 3.1-3.3, respectively.

**BNN reconstruction**

Following the input data conversion method in Section 3.1, the raw data of Cifar-10 dataset can be denoted as $Cifar_{(32,32,3)}$. Each pixel value is represented by a non-negative integer with magnitude $A=255$. Thus, $N=\text{ceil}(\log_2(255+1))=8$ bits are enough for lossless binary representation. Then, the bit-sliced input can be denoted as $Cifar^b_{(32,32,24)}$.

As illustrated in Section 3.1, the proposed structure of the reconstructed BNN is different from the original BNN in both input format and the first layer. Table 1 compares the performance results of three network structure with different input and 1st layer. The baseline BNN design is the one in [Hubara *et al.*, 2016], with full precision input and a binarized 1st layer. Here we define a CNN with bit-sliced input, binarized weights and activations as FBNN. FBNN has bit slices input but BNN does not. By training with the method in Section 3.1, FBNN shows 2.4% in the accuracy drop, compared with BNN. The accuracy here is affected by computational complexity degradation in the 1st layer and unnormalized input data. It also gives us some insights that The FBNN is hard to get a good accuracy rate, which is in accord with Tang et al.'s opinion in [Tang *et al.*, 2017]. By introducing bit slices input and non-binary 1st layer to reconstruct the BNN (as we proposed in Section 3.1), the accuracy drop can be compensated as shown in Table 1. We can even get a better error rate than the baseline BNN with a slightly increased network size. It also gives more margin in compressing the network.

Table 1. Performance comparison with different input format and 1st layer configuration

| Arch. | Input | First layer | Network size | Error rate |
|---|---|---|---|---|
| BNN | full precision | binary | 1x | 11.6% |
| FBNN | bit slices | binary | 1.01x | 14.0% |
| Reconstructed BNN | bit slices | non-binary | 1.1x | 10.1% |

Table 2. Sensitivity analysis of single bit slice in each channel with random noise injected

| Arch. | $N^{th}$ bit | ERR/% | ΔERR/% | Arch. | $N^{th}$ bit | ERR/% | ΔERR/% |
|---|---|---|---|---|---|---|---|
| BNN | 0 | 11.6 | 0.0 | FNN | 0 | 10.4 | 0.0 |
| Recon. BNN | 0 | 10.1 | -1.5 | FNN | 0 | 10.4 | 0.0 |
|  | 1 | 9.8 | -1.9 |  | 1 | 10.4 | 0.0 |
|  | 2 | 10.0 | -1.6 |  | 2 | 10.4 | 0.1 |
|  | 3 | 10.1 | -1.6 |  | 3 | 10.4 | 0.1 |
|  | **4** | **10.5** | **-1.2** |  | **4** | **10.9** | **0.5** |
|  | 5 | 12.5 | 0.8 |  | 5 | 13.0 | 2.6 |
|  | 6 | 20.9 | 9.2 |  | 6 | 21.4 | 11.1 |
|  | 7 | 40.3 | 28.6 |  | 7 | 43.8 | 33.4 |

Table 3. Sensitivity analysis of 1-$N^{th}$ multiple bit slices in each channel with random noise injected

| Arch. | 1-$N^{th}$ bits | ERR/% | ΔERR/% | Arch. | 1-$N^{th}$ bits | ERR/% | ΔERR/% |
|---|---|---|---|---|---|---|---|
| BNN | 0 | 11.6 | 0.0 | FNN | 0 | 10.4 | 0.0 |
| Recon. BNN | 0 | 10.1 | -1.5 | FNN | 0 | 10.4 | 0.0 |
|  | 1 | 9.8 | -1.9 |  | 1 | 10.4 | 0.1 |
|  | 1-2 | 9.9 | -1.7 |  | 1-2 | 10.5 | 0.2 |
|  | 1-3 | 9.9 | -1.8 |  | 1-3 | 10.5 | 0.2 |
|  | **1-4** | **10.7** | **-0.9** |  | **1-4** | **11.3** | **1.0** |
|  | 1-5 | 13.6 | 1.9 |  | 1-5 | 14.4 | 4.1 |
|  | 1-6 | 24.3 | 12.6 |  | 1-6 | 23.3 | 13.0 |
|  | 1-7 | 46.1 | 34.5 |  | 1-7 | 54.1 | 43.7 |

Table 4. Performance of CBNNs on Cifar-10

| Arch. | 1-$N^{th}$ bits | ERR % | ΔERR % | Network size | | GOPs | |
|---|---|---|---|---|---|---|---|
| | | | | MB | CP. ratio | # | CP. ratio |
| BNN | 0 | 11.6 | 0.0 | 1.75 | 1x | 1.23 | 1x |
| CBNN | 1 | 10.3 | -1.3 | 1.38 | 1.3x | 0.98 | 1.3x |
|  | 2 | 10.6 | -1.0 | 1.02 | 1.7x | 0.72 | 1.7x |
|  | 3 | 10.8 | -0.8 | 0.71 | 2.5x | 0.50 | 2.5x |
|  | **4** | **11.8** | **0.2** | **0.45** | **3.9x** | **0.32** | **3.8x** |
|  | 5 | 14.2 | 2.6 | 0.25 | 7.0x | 0.18 | 6.8x |

**Sensitivity analysis of the reconstructed BNN**
With a pre-trained reconstructed BNN presented in the last section, now we can do bit-level sensitivity analysis as stated in Section 3.2.

First, we analyze the sensitivity of a single bit slice. The results are shown in Table 2. The data shows in Table 2 is the average over 10 trials. In addition to the reconstructed BNN, we also evaluate the bit-level sensitivity of the input with its full-precision counterpart, which is denoted as FNN. With FNN, we intend to show that the data itself has redundancy, which can be reflected in both binary domain or fixed-point domain with the same pattern. We take the architecture in the first row as the reference design. The 1st row of *ERR* column is the $ERR_{ref}$ and the others are $ERR_{inf}$. $ΔERR=ERR_{inf}- ERR_{ref}$. BNN is the reference design for the reconstructed BNN. FNN with non-distorted input is the reference design for the full-precision ones. It is interesting that the 1st, 2nd and 3rd bit slices are at the same sensitivity level, concluded from the almost unchanged *ΔERR*. We define the turning point of error in sensitivity analysis as the point where *ΔERR* flips the sign or increases abruptly. The turning point here is the 5th bit.

Second, we analyze the sensitivity of bit slices stacks. Each stack contains 1st to $N^{th}$ bit slices in each color channel. The results are shown in Table 3. For the 1st, 2nd and 3rd bit slices, it makes no difference if distortion is injected in one of them or all of them. The 4th makes a slight difference of around 0.5% accuracy drop and the 5th bit is also the turning point with around 3% accuracy drop.

Even when we randomize 50% of the entire input values (1st to 4th bit slices) and the variation propagates through the entire network, the accuracy doesn't change much. Therefore, these bits are useless in the training stage. This validates the hypothesis that the BNN still has redundancy. In Fig. 4, the error rate turning point is circled at the 5th bit slice. The trend of error rate in the binary domain and full-precision domain (shown in Fig. 4) align well. In order to make the entire process be automatic, we can simple set an error-tolerant threshold $ERR_{th}$ to determine how many bits are prunable. Here, $ERR_{th}$ is set to 1%. We can conclude that 1st-4th bit slices here are redundant and prunable through bit-level sensitivity analysis. Accordingly, the reconstructed BNN can be shrunk to reduce the redundancy and get a more compact architecture.

**Rebuild a compact BNN (CBNN)**
Since 4 out of 8 bit slices are prunable, we can rebuild a compact BNN with the depth of each layer shrunk by half. The performance of CBNN is shown in Table 4. CP. Ratio represents compression ratio and GOPs stands for Giga operations (one operation is either an addition or a multiplication).

Regarding the network size, we use 16 bits for measuring non-binary weights in the 1st layer, since it has been proved that 16 bit is enough to maintain the same accuracy [Suda *et*

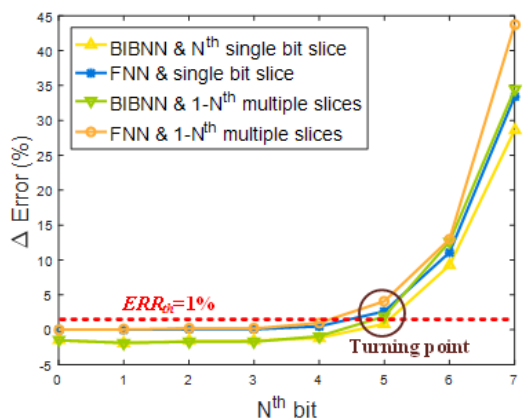

Figure 4. Error rate of randomizing one or multiple bit slices in sensitivity analysis

Table 5. Performance results of CBNNs on SVHN/Chars47k/GTSRB datasets

| Dataset | 1-N[th] bits | ERR % | ΔERR % | Network size MB | Network size CP. ratio | GOPs # | GOPs CP. ratio |
|---|---|---|---|---|---|---|---|
| SVHN | 0 | 4.8 | 0.0 | 0.44 | 1x | 0.31 | 1x |
| | 1 | 4.9 | 0.1 | 0.36 | 1.2x | 0.26 | 1.2x |
| | 2 | 5.1 | 0.3 | 0.26 | 1.7x | 0.19 | 1.6x |
| | **3** | **5.0** | **0.2** | **0.18** | **2.4x** | **0.13** | **2.4x** |
| | 4 | 6.6 | 1.8 | 0.12 | 3.7x | 0.08 | 3.7x |
| Chars47k | 0 | 15.4 | 0.0 | 0.44 | 1x | 0.31 | 1x |
| | 1 | 15.3 | -0.1 | 0.36 | 1.2x | 0.26 | 1.2x |
| | 2 | 15.3 | -0.1 | 0.26 | 1.7x | 0.19 | 1.6x |
| | 3 | 15.2 | -0.2 | 0.18 | 2.4x | 0.13 | 2.4x |
| | **4** | **16.3** | **1.0** | **0.12** | **3.7x** | **0.08** | **3.7x** |
| GTSRB | 0 | 1.0 | 0.0 | 1.81 | 1x | 3.89 | 1x |
| | 1 | 1.0 | 0.0 | 1.39 | 1.3x | 2.98 | 1.3x |
| | 2 | 1.2 | 0.2 | 1.02 | 1.8x | 2.19 | 1.8x |
| | 3 | 1.6 | 0.6 | 0.71 | 2.5x | 1.52 | 2.6x |
| | **4** | **2.0** | **1.0** | **0.46** | **3.9x** | **0.97** | **4.0x** |

*al.*, 2016]. We also show the alternatives of pruning 1-$N^{th}$ ($N = 1, 2, ..., 5$) bit slices and shrink the layerwise depth by 1/8 to 5/8. The results align with the sensitivity analysis that 1-$3^{rd}$ bit slices have little impact on the classification performance. The choice of pruning 1-$4^{th}$ bit slices is the optimal one to maximize the compression ratio with <1% accuracy drop. Since the size of the $1^{st}$ layer is larger than that of BNN, we cannot achieve the ideal network size compress ratio (4x) regarding the entire network. The actual compression ratio of the network size is 3.9x and the compression ratio of number of GOPs is 3.8x.

### 4.2 Experiment on SVHN/Chars74K/GTSRB datasets

In this section, we will skip the sensitivity analysis and just show the result comparison between the baseline and the final CBNNs we get in the same procedure.

For SVHN and Char74K datasets, we use a baseline architecture that has half of the depth in each layer as the one for Cifar-10. For GTSRB, we use a baseline architecture that has the same filter configuration as the one for Cifar-10. Since the input size of GTSRB is larger than Cifar-10, so the network for GTSRB has the same depth but larger width and height in each layer.

In Table 5, it shows the performance results of CBNNs evaluating on different datasets and network setting. The baseline for each dataset is shown in the first row of each dataset region. For Chars47k and GTSRB, the CBNNs are able to maintain no more than 1% accuracy drop, achieving 3.7x and 3.9x network size reduction, respectively. For SVHN dataset, the accuracy drop between pruning 1-$3^{rd}$ bits and pruning 1-$4^{th}$ bits is large. In order to preserve no more than 1% accuracy drop, the network size reduction is yield to 2.4x.

### 4.3 Runtime evaluation

We evaluate the actual runtime performance of CBNNs by Nvidia GPU Titan X. The batch size is fixed as 128 in all the experiments. We use the same XNOR-based GPU kernel as [Hubara *et al.*, 2016] for CBNN implementation. The computational time is calculated by averaging over 10 runs.

Fig. 5 (a) illustrates the actual runtime and runtime speedup of 4 CBNNs compared with their baseline BNNs. The configurations are the same as the highlight ones in Table 4 and Table 5. For the CBNNs processing Cifar10, GTSRB and Char47k datasets, their network size and total GOPs shrink 3.7-4.0x, resulting in the speedup of 1.7-2.0x. For the CBNN processing the SVHN dataset, its network size and total GOPs shrinks 2.4x, resulting in a speedup of 1.4x. In Fig. 5 (b), we normalize the runtime performance to a full-precision CNN (FNN). It is proved in [Han *et al.*, 2016], combining pruning, quantization and Huffman coding technique, an FNN can achieve up to 4x speedup, which is shown in the green bar. Hubara et al. demonstrate that a multilayer perceptron BNN can get 5x speedup compared with its full-precision counterpart. On top of the BNN, the proposed CBNN can give extra 1.4-2.0x speedup. Therefore, the CBNN can achieve 7.0-9.9x speedup compared with FNN.

## 5 Conclusion

In this paper, we propose a novel flow to explore the redundancy of BNN and remove the redundancy by bit-level sensitivity analysis and data pruning. In order to build a compact BNN, one should follow these three steps. Specifically, first reconstruct a BNN with bit-sliced input and non-binary 1st layer. Then, inject randomly binarized bit slices to analyze the sensitivity level of each bit slice to the classification error rate. After that, prune *P* accuracy insensitive bit slices out of total *N* slices and rebuild a CBNN with depth shrunk by (*N/P*) times in each layer. The experiment results show that the error variation trend in sensitivity analysis of the reconstructed BNN is well aligned with that of CBNN. In addition, the CBNN is able to get 2.4-3.9x network compression ratio and 2.4-4.0x computational complexity reduction (in terms of GOPs) with no more than 1% accuracy loss compared with BNN. The actual runtime can be reduced by 1.4-2x and 7.0-9.9x compared with the baseline BNN and its full-precision counterpart, respectively.

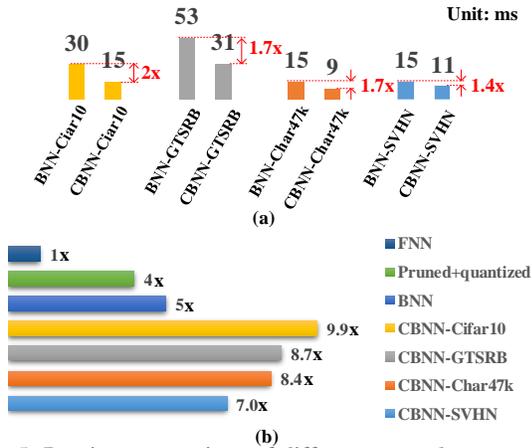

Figure 5. Runtime comparison of different network compression technique